\documentclass[letterpaper, 10 pt, journal, twoside]{IEEEtran}
\usepackage[nolist]{acronym} 
\usepackage{amsfonts,siunitx,booktabs,cite} 
\usepackage{graphics} 
\usepackage{epsfig} 
\usepackage{mathptmx} 
\usepackage{times} 
\usepackage{amsmath} 
\usepackage{amssymb}  
\usepackage[linesnumbered,ruled]{algorithm2e}
\usepackage{stmaryrd}
\usepackage{soul}
\usepackage{bm}

\usepackage{graphicx}
\usepackage{caption}
\let\labelindent\relax
\usepackage{enumitem}
\usepackage{svg}
\usepackage{float}
\usepackage[normalem]{ulem}
\usepackage{hyperref}
\usepackage{algpseudocode}

\usepackage{scalerel}
\usepackage{tikz,standalone,pgfplots}

\DeclareMathAlphabet{\pazocal}{OMS}{zplm}{m}{n}
\newcommand{\unif}{\pazocal{U}}

\newacro{dnn}[DNN]{Deep Neural Network}
\newacro{fcn}[FCN]{Fully Convolutional Network}
\newacro{sdf}[SDF]{Signed Distance Function}
\newacro{cnn}[CNN]{Convolutional Neural Network}
\newacro{gnn}[GNN]{Graph Neural Network}
\newacro{dl}[DL]{Deep Learning}
\newacro{ml}[ML]{Machine Learning}
\newacro{mc}[MC]{Monte Carlo}
\newacro{mlp}[MLP]{Multi-Layer Perceptron}
\newacro{dof}[DoF]{Degrees of Freedom}
\newacro{vae}[VAE]{Variational Autoencoder}
\newacro{cvae}[CVAE]{Conditional Variational Autoencoder}
\newacro{methodname}[VCGS]{Variational Constrained Grasp Sampler}
\newacro{fps}[FPS]{Farthest Point Sampling}
\newacro{tai}[TaI]{Target as Input}
\newacro{pca}[PCA]{Principal Component Analysis}
\newacro{pc}[PC]{Principal Component}
\newacro{auc}[AUC]{Area Under the Curve}
\newacro{elbo}[ELBO]{Evidence Lower Bound}

\newcommand{\equationref}[1]{\hyperref[#1]{Eq.~\ref*{#1}}}
\newcommand{\figref}[1]{\hyperref[#1]{Fig.~\ref*{#1}}}
\newcommand{\tabref}[1]{\hyperref[#1]{Table~\ref*{#1}}}
\newcommand{\secref}[1]{\hyperref[#1]{Section~\ref*{#1}}}
\newcommand{\algoref}[1]{\hyperref[#1]{Algorithm~\ref*{#1}}}

\newcommand{\norm}[1]{\left\lVert#1\right\rVert}

\newcommand{\matr}[1]{\mathbf{#1}}

\newcommand*{\prob}{\mathsf{P}}

\newcommand{\etal}[1]{#1 et al.}

\def\oldmethodname{GoNet}
\def\methodname{CAPGrasp}

\def\graspnet{GraspNet}

\def\gonet{GoNet}
\def\gpd{GPD}
\def\edgegrasp{EdgeGraspNet}
\def\equiv{$\mathbb{R}^3\times \text{SO(2)-equivariant}$}

\def\sota{state-of-the-art}
\def\ie{, \textit{i.e.}, }

\def\pointnet{PointNet++~\cite{qiPointNetDeepHierarchical2017a}}

\ifCLASSINFOpdf
\else
\fi
\begin{document}
%
\title{CAPGrasp: An \equiv{} Continuous Approach-Constrained Generative Grasp Sampler}
%
%
%


\author{Zehang Weng, Haofei Lu, Jens Lundell, and Danica Kragic%
\thanks{All authors are with the division of Robotics, Perception, and Learning (RPL) at KTH, Stockholm, Sweden. 
        {\tt\footnotesize \{zehang,haofeil,jelundel,dani\}@kth.se}}%
}
\maketitle

\begin{abstract}
We propose \methodname{}, an \equiv{} 6-\ac{dof} continuous approach-constrained generative grasp sampler. It includes a novel learning strategy for training \methodname{} that eliminates the need to curate massive conditionally labeled datasets and a constrained grasp refinement technique that improves grasp poses while respecting the grasp approach directional constraints. The experimental results demonstrate that \methodname{} is more than three times as sample efficient as unconstrained grasp samplers while achieving up to 38\% grasp success rate improvement. \methodname{} also achieves 4-10\% higher grasp success rates than constrained but noncontinuous 
 grasp samplers. Overall, \methodname{} is a sample-efficient solution when grasps must originate from specific directions, such as grasping in confined spaces. Videos and code are publicly accessible at \url{https://wengzehang.github.io/CAPGrasp/}.
\end{abstract}


%
\IEEEpeerreviewmaketitle


\section{Introduction}

\IEEEPARstart{W}e address the problem of synthesizing 6-\ac{dof} grasps whose approach directions are within a specific target range.
This problem, known as approach-constrained grasp synthesis \cite{weng2023gonet}, is present in many robotics tasks, including interactive tactile sensing \cite{welle2023enabling}, human-to-robot handover \cite{yang2021reactive} and task-based grasping \cite{kokic2020learning}. Approach-constrained grasp synthesis can be solved by i) generating grasps around the target object and eliminating those not adhering to the specified constraints or by ii) generating grasps that meet the specified approach constraints. In our previous work \cite{weng2023gonet}, where we introduced a 6-\ac{dof} \textit{discrete} approach-constrained generative grasp sampler, we demonstrated that methods following ii) are more accurate and efficient at generating grasps at the target area than methods following i). Unfortunately, the discreteness of the grasp sampler in \cite{weng2023gonet} restricted the possible approach directions to be within a predefined discrete subset of SO(3), meaning that it could only generate grasps from specific directions, which left the problem of generating grasps with \textit{any} approach direction
in SO(3) unsolved.

\begin{figure}[ht]
    \centering
    \includegraphics[width=0.65\linewidth]{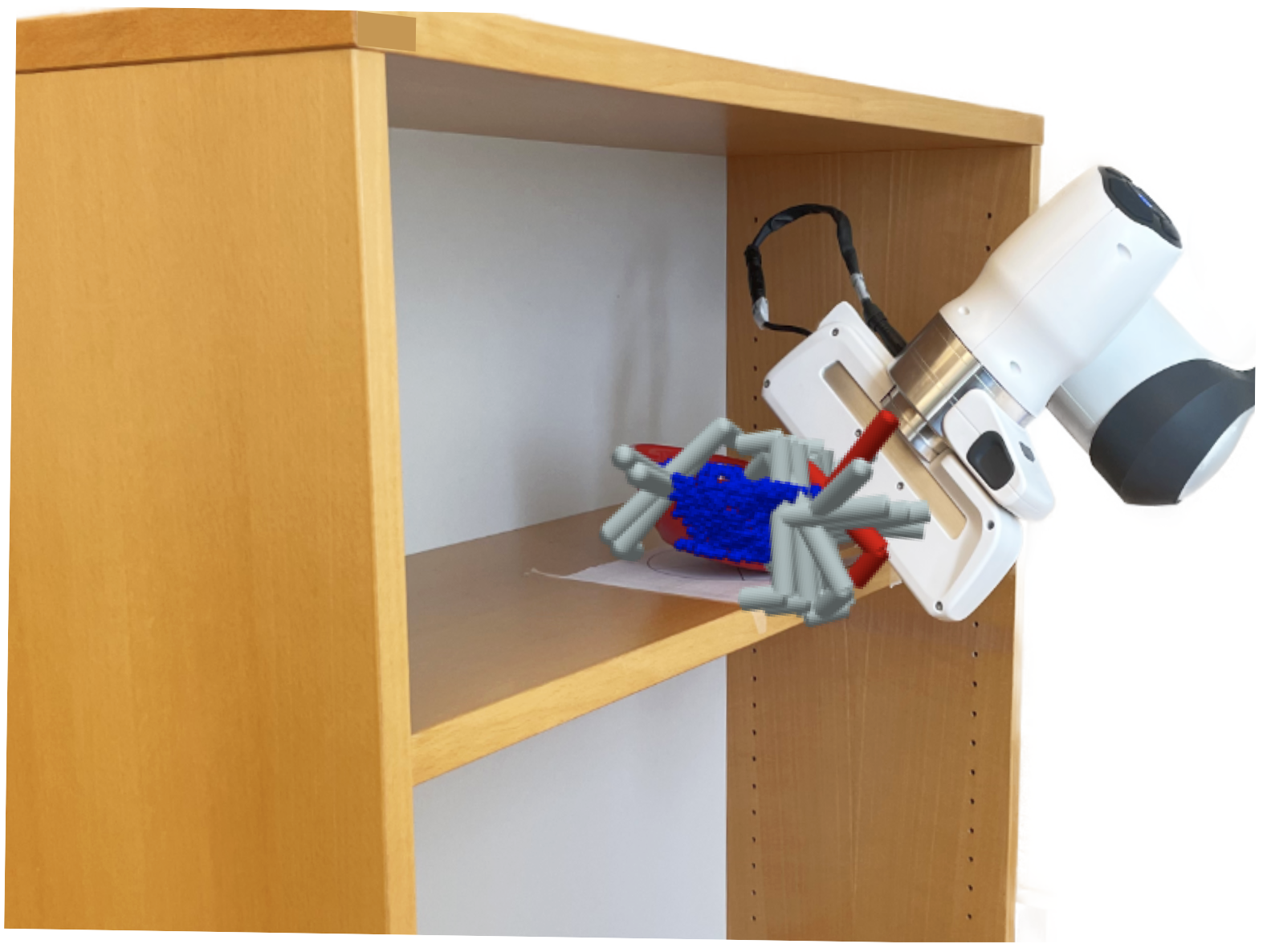}
    \caption{An example of picking from a shelf: generating grasps with specified approach constraints and choosing the highest scoring one (red) for execution.}
    \label{fig:confine_space_bowl}
\end{figure}

In this work, we address that specific problem with the 6-\ac{dof} \textit{continuous} approach-constrained generative grasp sampler called \methodname{}. Given an object point cloud and approach constraint in SO(3), \methodname{} samples a diverse set of grasps that adheres to the constraint as shown in \figref{fig:confine_space_bowl}. Interstingly, \methodname{} can be made \equiv{} as the continuous approach-constrained formulation reveals symmetries in the input data, something the discrete formulation cannot achieve. In addition to \methodname{}, we also introduce a new training algorithm that eliminates the need for curating extensive training data and a constrained grasp refinement approach that improves the generated grasps in terms of success rate without violating the approach constraint.

In summary, our main contributions are:
\begin{itemize}
\item \methodname{}: A 6-\ac{dof} \equiv{} continuous approach-constrained generative grasp sampler (\secref{sec:\methodname{}});
\item A novel training algorithm that eliminates the need for curating extensive offline datasets (\secref{sec:training\methodname{}});
\item A grasp refinement method that refines grasps abiding by the constraint (\secref{sec:refinement\methodname{}}).
\end{itemize}

We benchmarked \methodname{} against three learning-based methods: a discrete approach-constrained grasp sampler \gonet{} \cite{weng2023gonet}, an unconstrained grasp sampler \graspnet{} \cite{mousavian20196}, and an SE(3)-equivariant unconstrained grasp sampler \edgegrasp{} \cite{huang2023edge}. Additionally, we compared \methodname{} to the non-learning-based unconstrained sampler \gpd{} \cite{ten2017grasp}. In total, we evaluated 17.6 million simulated grasps and 450 grasps in the real world. All results show that \methodname{} surpasses the baselines in both efficiency and grasp success rate.

\section{Related Work}
\label{sec:related_work}

This work spans three topics: constrained grasp sampling, equivariance in robotic learning, and continuous conditional generative models. Next, we will review these topics from both the robotics and the machine learning perspective.

\subsection{Constrained Grasp Sampling}

Constrained grasp sampling research primarily falls under three categories: 4-\ac{dof} grasping, task-based grasping, and geometrically constrained grasping. In 4-\ac{dof} grasping, the approach direction is limited to be perpendicular to a specific plane, facilitating learning by reducing the potential grasp space \cite{morrisonClosingLoopRobotic2018b,mahlerDexNetDeepLearning2017a,zhou2018fully,satish2019policy,kumra2020antipodal, wang2022robot}. Task-based grasping involves sampling grasps for specific tasks, either using optimization-based \cite{borst2004grasp,haschke2005task} or learning-based methods \cite{murali2021same,antonovaGlobalSearchBernoulli2018a,kokic2017affordance,song2010learning,song2015task,fang2020learning,detry2017task}. Geometrically constrained grasping, on the other hand, aligns the grasp with the object model's geometry \cite{pas2016localizing,stoyanov2016grasp,balasubramanian2012physical}. 

None of the aforementioned approaches can constrain the generated grasps to \textit{any} subset of SE(3). However, two recent methods have been proposed to address this issue: one that explicitly constrains the position of the grasp \cite{lundell2023constrained} and another that constrains the approach direction \cite{weng2023gonet}. Similar to \cite{weng2023gonet}, we also consider approach-constrained grasp sampling, but instead of discretizing the whole SO(3), which requires training different networks to represent different approach directions, we propose a continuously constrained model that can represent all types of discretizations with \textit{one} model.   

\subsection{Equivariance in Robotic Learning}

In machine learning, equivariance refers to a property where the transformation of an input leads to an equivalent output transformation, allowing the model to recognize patterns irrespective of their orientation or location in the input space. There are two prevailing options for ensuring equivariance: 1) make the model equivariant or 2) make the data equivariant. Prior works on equivariant robotic grasp learning \cite{wang2022robot, huang2023edge} have focused on equivariant models. In contrast, this work proposes a solution for making the data equivariant.

\subsection{Continuous Conditional Generative Models}

Conditional generative models have been extensively researched in computer vision \cite{wu2017survey,mirza2014conditional,isola2017image}. The main challenge with conditional models is the time-consuming process of curating training data, as each input sample requires a conditional label. Consequently, the conditional labels are primarily discrete and finite \cite{wu2017survey,mirza2014conditional,isola2017image}. Only recently,  \etal{Ding} \cite{ding2022continuous} proposed a continuously conditioned generative image synthesis network by perturbing discrete labels to generate continuous conditions via a vicinity loss. Unfortunately, that approach is impractical for our target problem as it would require dense grasp sampling on the whole SE(3). Instead, we propose a new efficient training algorithm that generates conditional labels on the fly while training.

\section{Problem Statement}
\label{sec:problem_statement}

\begin{figure}[ht]
    \centering
    \vspace{9pt}
    \includegraphics[width=0.65\linewidth]{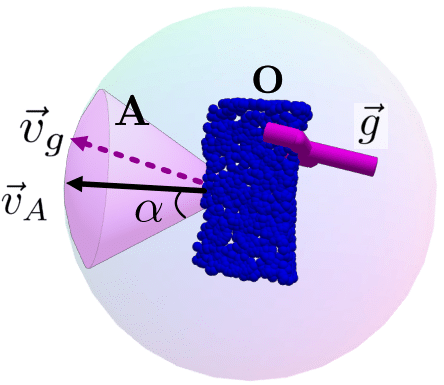}
    \caption{A visualization of the problem statement. $\matr{O}$ is represented by the blue point cloud, while $\matr{A}$, parameterized by $\Vec{v}_{A}$ and $\alpha$, is visualized as the purple cone. The goal is to generate a grasp pose $\Vec{g}$ whose unit approach vector $\Vec{v}_{g}$ lies inside $\matr{A}$.}
    \label{fig:problem_statement}
    \vspace{-9pt}
\end{figure}

 Given an object point cloud $\matr{O}\in \mathbb{R}^{\text{N}\times 3}$ and a spherical cone $\matr{A}$ parameterized by a unit directional vector $\Vec{v}_A$ and a maximum allowed half-angle $\alpha$, the objective is to generate a set of successful grasp poses $\matr{G}$ on the object such that the approach direction is within $\matr{A}$ (see \figref{fig:problem_statement}). We represent a grasp pose $ \vec{g} \in \matr{G} = [\Vec{q}, \Vec{p}] \in \mathbb{R}^7$ by a unit quaternion $ \Vec{q} \in \mathbb{R}^4$ and a 3-D position  $\Vec{p} \in \mathbb{R}^3$. We let $\matr{A}$ constrain the grasp approach direction while leaving the remaining \acp{dof}\ie{} the roll angle around the approach direction and the gripper's position, unconstrained. 

Mathematically, the objective is to learn the joint distribution $\prob{(\Vec{g} \in \matr{G}, \text{S}=1 | \matr{O}, \matr{A}=[\alpha,\Vec{v}_A])}$, where S is a binary variable representing grasp success (S=1) or failure (S=0). We can further factorize $\prob{(\Vec{g} \in \matr{G}, \text{S}=1 | \matr{O}, \matr{A}=[\alpha,\Vec{v}_A])}$ into $\prob{(\text{S}=1 | \Vec{g} \in \matr{G}, \matr{O})}\prob{(\Vec{g} \in \matr{G}|\matr{O}, \matr{A}=[\alpha,\Vec{v}_A])}$ because S is conditionally independent of $\matr{A}$ given a grasp pose $\Vec{g}$. Henceforth, $\prob{(\text{S}=1 | \Vec{g} \in \matr{G}, \matr{O})}$ is called the grasp discriminator and $\prob{(\Vec{g} \in \matr{G}|\matr{O}, \matr{A}=[\alpha,\Vec{v}_A])}$ the grasp generator. We approximate these two distributions by separate \acp{dnn}: $\mathcal{Q}_{\boldsymbol{\theta}}(\Vec{g} \in \matr{G}|\matr{O}, \matr{A}=[\alpha,\Vec{v}_A])\approx \prob{(\Vec{g} \in \matr{G}|\matr{O}, \matr{A}=[\alpha,\Vec{v}_A])}$ and $\mathcal{D}_{\boldsymbol{\psi}}(\text{S}=1 | \Vec{g} \in \matr{G}, \matr{O}) \approx \prob{(\text{S}=1 | \Vec{g} \in \matr{G}, \matr{O})}$, with learnable parameters $\boldsymbol{\theta}$ and $\boldsymbol{\psi}$. The objective then becomes learning $\boldsymbol{\theta}$ and $\boldsymbol{\psi}$ from data.

\section{Method}

As mentioned in \secref{sec:problem_statement}, the objective is to learn the parameters for the generative grasp sampler $\mathcal{Q}_{\boldsymbol{\theta}}$ and the grasp discriminator $\mathcal{D}_{\boldsymbol{\psi}}$ from data. The following subsections detail \methodname{}, our continuously constrained \equiv{} generative grasp sampler, the novel learning strategy for training \methodname{}, the grasp discriminator, the constrained grasp refinement approach, and implementation details.

\subsection{\methodname{}}
\label{sec:\methodname{}}

\begin{figure}[ht]
    \centering
    \vspace{9pt}
        \includegraphics[width=0.75\linewidth]{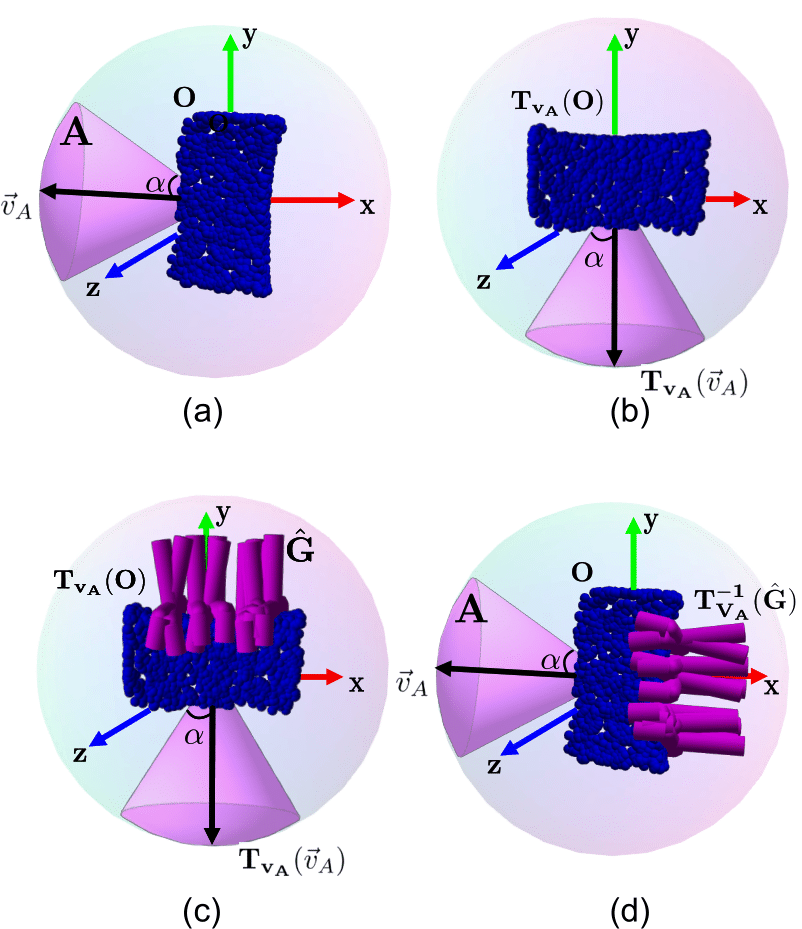}
    \caption{\methodname{} grasp sampling. (a) $\matr{O}$ and $\matr{A}$ are in the camera space. (b) $\matr{O}$ and $\matr{A}$ are transformed into the approach space using the transformation $\matr{T_{v_{A}}}$ that aligns $\Vec{v}_A$ with $-\Vec{y}$. (c) \methodname{} generates grasps $\matr{\hat{G}}$ whose approach directions are within $\alpha$. (d) $\matr{\hat{G}}$ is transformed back to the camera space using $\matr{T_{v_{A}}^{-1}}$.}
    \label{fig:gonet_x_approachspace}
    \vspace{-1pt}
\end{figure}

We propose the \ac{cvae} \methodname{} for learning $\mathcal{Q}_{\boldsymbol{\theta}}(\Vec{g} \in \matr{G}|\matr{O}, \matr{A}=[\alpha,\Vec{v}_A])$. Interestingly, as depicted in \figref{fig:gonet_x_approachspace}, it is enough to condition $\mathcal{Q}_{\boldsymbol{\theta}}$ on $\matr{O}$ and $\alpha$ as long as we transform $\matr{O}$ and $\matr{G}$ to and from a new unique coordinate system that aligns $\Vec{v}_{A}$ with the negative y-axis of the camera coordinate system $-\Vec{y}$. In this new coordinate system, henceforth referred to as the  \textit{approach space}, grasps are generated from a top-down direction, eliminating the need to explicitly condition on $\Vec{v}_A$.

To transform to and from the approach space, we need to find a unique rigid transformation matrix 
\begin{align}
\label{eq:approach_space_transformation}
\matr{T_{v_{A}}} = \begin{pmatrix}
\matr{R_{v_{A}}} & \textbf{0}\\
\textbf{0}^T & 1 
\end{pmatrix}.
\end{align}
However, an infinite number of such transformations exist. Thus, to find a unique $\matr{T_{v_{A}}}$, we restrict the rotation matrix $\matr{R_{v_{A}}}$ to the plane spanned by $\Vec{v}_{A}$ and $-\vec{y}$. Using this restriction, $\matr{R_{v_{A}}}$ is determined using Rodrigues' rotation formula
\begin{align}
\label{eq:approach_space_transformation_rod}
\matr{R_{v_{A}}} = \matr{I} + \sin{(\phi)}\widehat{\Vec{v}_{r}} + (1-\cos{(\phi)})(\widehat{\Vec{v}_{r}})^2,
\end{align}
where $\phi=\arccos{(\langle\Vec{v}_{A}, -\Vec{y}\rangle)}$, $\Vec{v_{r}}=\Vec{v}_{A} \times \Vec{y}$, and $\widehat{\Vec{v}_{r}}$ is the $\mathbb{R}^{3\times 3}$ skew-symmetric matrix of $\Vec{v}_{r}$. Using the new approach space, $\mathcal{Q}_{\boldsymbol{\theta}}(\matr{G}|\matr{O}, \matr{A})$ turns into $\matr{T_{v_{A}}}^{-1}[\mathcal{Q}_{\boldsymbol{\theta}}(\matr{T_{v_{A}}}(\matr{G})|\matr{T_{v_{A}}}(\matr{O}), \alpha)]$.

The approach space offers two advantages. Firstly, this space eliminates the need to optimize the \ac{cvae} with explicit yaw and pitch labels, simplifying the training phase. Secondly, the approach space makes \methodname{} \equiv{}, providing consistent feature representation regardless of variance in translation or rotation. The translation part of the \equiv{} is due to $\matr{O}$ always being zero-centered, while the rotation part stems from canonicalizing $\matr{A}$ and $\matr{O}$ by rotating within a unique 2D plane.

As \methodname{} is a \ac{cvae} it consists of an encoder $q_{\boldsymbol{\zeta}}(\mathbf{z}\mid\matr{T_{v_{A}}}(\matr{O}),\matr{T_{v_{A}}}(\matr{G}),\alpha)$ and a decoder $p_{\boldsymbol{\chi}}(\matr{G}\mid\matr{T_{v_{A}}}(\matr{O}),\mathbf{z},\alpha)$, where $\mathbf{z}\in \mathbb{R}^{\text{L}}$ are latent vectors sampled from a zero-mean Gaussian distribution. The backbone architecture of \methodname{} is \pointnet{}, which requires point clouds as input. The input point cloud to the encoder is $\matr{O}$ with $\Vec{g} \in \matr{G}$ and $\alpha$ as additional point-wise features. The decoder takes the same input as the encoder but swaps the point-wise features $\Vec{g} \in \matr{G}$ for $\mathbf{z}$.   

\methodname{} is trained using the standard \ac{elbo} loss:
\begin{align}
\label{eq:vae_loss}
    \mathcal{L}_{\text{VAE}} = \mathcal{L}_{r}(\matr{G}^*,\hat{\matr{G}}) + \beta \mathcal{L}_{\text{KL}}[q_{\boldsymbol{\zeta}}(\mathbf{z}\mid\matr{T_{v_{A}}}(\matr{O}),\matr{G}^*,\alpha),~\mathcal{N}(\matr{0},\matr{I})],
\end{align}
where $\mathcal{L}_{r}$ is the reconstruction loss, $\mathcal{L_{\text{KL}}}$ is the KL-divergence loss, $\beta$ is the KL-divergence loss weight, $\matr{\hat{G}}$ are the generated grasps, and $\matr{G}^*$ are the ground-truth grasps in the approach space. The reconstruction loss $\mathcal{L}_r(\matr{G}^*,\hat{\matr{G}})$ is defined as
\begin{align}
    \mathcal{L}_r(\matr{G}^*,\hat{\matr{G}}) = \norm{\text{h}(\matr{G}^*)-\text{h}(\mathbf{\hat{\matr{G}}})}_1,
    \label{eq:rec_loss}
\end{align}
where $\text{h}: \mathbb{R}^{7} \rightarrow  \mathbb{R}^{6\times 3}$ maps the 7-D grasp pose $\Vec{g} \in \matr{G}$ to a coarse point cloud representation of the grasp. The reason for using this loss function is that it jointly penalizes position and orientation errors \cite{mousavian20196}.  

To generate M grasps $\matr{G}$ on $\matr{O}$ while respecting $\matr{A}$, we first calculate $\matr{T_{v_{A}}}$ and use it to transform $\matr{O}$ to the approach space. Then we sample M random latent vectors $\mathbf{z}_{1,~\dots,~\text{M}} \sim \mathcal{N}(\matr{0},\matr{I})$ and add each of them, together with $\alpha$, as additional point-wise features to an individual copy of $\matr{T_{v_{A}}}(\matr{O})$. Next, each copy is passed through $p_{\boldsymbol{\chi}}$ that generates $\matr{\hat{G}}$ in the approach space, which is finally transformed back into the camera coordinate system $\matr{G}=\matr{T_{v_{A}}}^{-1}(\matr{\hat{G}})$ before execution on the robot.

\subsection{Training \methodname{}}
\label{sec:training\methodname{}}

As mentioned in \secref{sec:related_work}, training \acp{cvae} requires large datasets as each input data requires a conditional label. This restriction is the primary reason training datasets for \acp{cvae} restrict the input and conditional values to be discrete. However, we cannot adhere to this restriction as the space of inputs $\matr{G}$ and conditions $\matr{A}$ is continuous. The same reasons prevent us from using the method for training continuous \acp{cvae} from \cite{ding2022continuous} as it requires starting from a dense but discrete set of $\matr{G}$ and $\matr{A}$ which is a prohibitively large space to cover in SE(3). Therefore, we propose the new efficient training scheme in \algoref{alg:training_alg} that labels $\matr{A}$ on-the-fly.

\begin{figure}[ht]
    \centering
    \vspace{9pt}
        \includegraphics[width=0.82\linewidth]{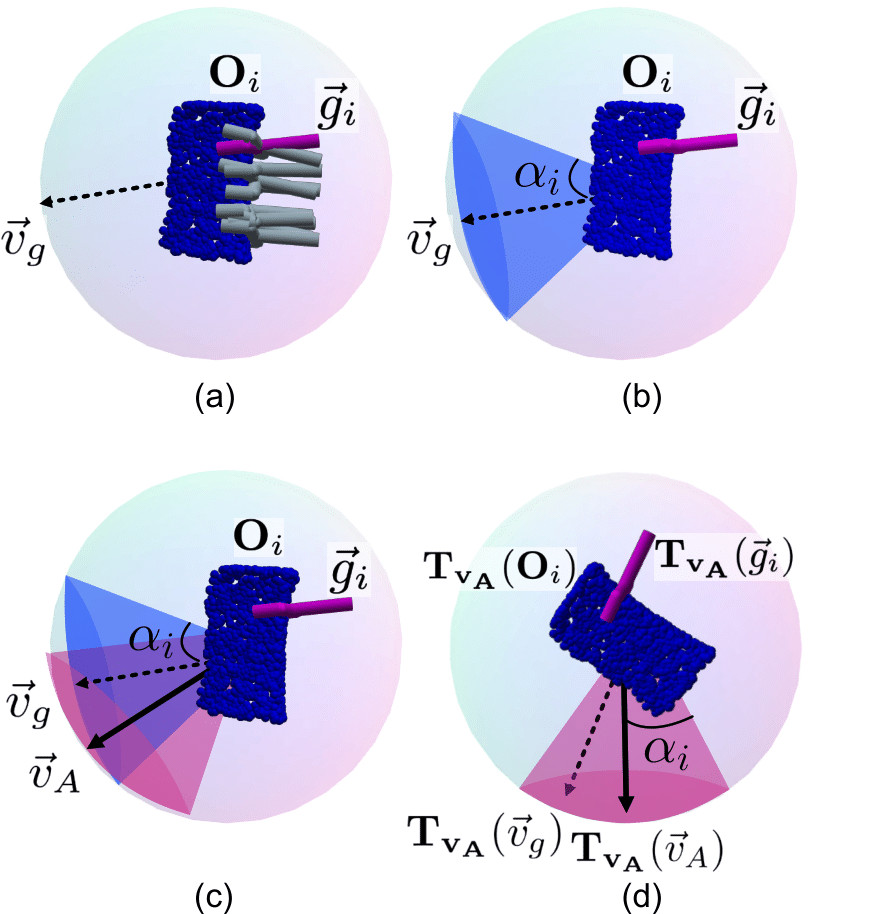}
    \caption{The training method for \methodname{}. (a) Sample an object-grasp pair in the camera space. (b) A random $\alpha_{i}$ is sampled (\textbf{Stage 1} in \algoref{alg:training_alg}). The resulting blue cone, parameterized by $\Vec{v}_{g}$ and $\alpha_{i}$, represents the space from which we can sample $\Vec{v}_A$. (c) $\Vec{v}_{A}$ is sampled inside the blue cone (\textbf{Stage 2} in \algoref{alg:training_alg}). The pink cone, parameterized by $\Vec{v}_{A}$ and $\alpha_{i}$, now represents $\matr{A}$. (d) The object-grasp pair is transformed into the approach space and becomes the conditional training pair.}
    \label{fig:learning_algorithm}
    \vspace{-2pt}
\end{figure}

 The key insight of the training algorithm, depicted in \figref{fig:learning_algorithm}, is that given a random $\Vec{g}\in \matr{G}$, it is possible to \textit{infer} an $\matr{A}$ that would label it. Specifically, for each $\Vec{g}\in \matr{G}$, we sample a random $\alpha_i$ (line \ref{line:stage_1}). Then, we sample a random $\Vec{v}_A$ that is inside the spherical cone parameterized by $\Vec{v}_g$ and $\alpha_i$  (line \ref{line:stage_2}). Next, we find $\matr{T_{v_{A}}}$ that rotates $\Vec{v}_A$ to the approach space (line \ref{line:transformation_matrix}), and finally construct the conditional training pair in the approach space (line \ref{line:construct_training_pair}). 

\begin{algorithm}[!ht]
    \caption{The training method for CAPGrasp.}\label{alg:training_alg}
    \footnotesize
    \SetAlgoLined
    \KwData{N object-grasp data pairs $\Omega^{r}=\{(\matr{O}_1^{r}, \Vec{g}_1^{r}),\dots,(\matr{O}_{\text{N}}^{r}, \Vec{g}_{\text{N}}^{r})\}$ in the camera space, a maximum half-angle $\alpha_{max}$, training batch size m, number of epochs K.}
    \KwResult{Optimized CAPGrasp.} 
    \For{$\text{k}=1$ \KwTo $\text{K}$}{

        Sample a subset of object-grasp pairs:  $\Omega^{s} = \{(\matr{O}_1, \Vec{g}_1),\dots,(\bm{O}_{m}, g_{m})\} \subset \Omega^{r}$.

        \For{ $(\matr{O}_{i}, \Vec{g}_{i})$ in $\Omega^{s}$} {
        
        Compute the unit approach vector $\Vec{v}_g$ of $\Vec{g}_{i}$.

        \textbf{Stage 1}: sample the conditional half-angle $\alpha_{i} \sim \unif (0,\alpha_{max}]$\label{line:stage_1}.



        \textbf{Stage 2:} sample a directional unit vector for the spherical cone $\Vec{v}_{A}$ such that $\arccos{(\langle\Vec{v}_{g},\Vec{v}_{A}\rangle)} \leq \alpha_{i}$\label{line:stage_2}.

        Compute the transformation matrix $\matr{T_{v_{A}}}$ using Eq.~\eqref{eq:approach_space_transformation}\label{line:transformation_matrix}.

        Construct the conditional training pair $\lambda_{i} = (\matr{T_{v_{A}}}(\matr{O}_i), \matr{T_{v_{A}}}(\Vec{g}_i), \alpha_{i})$ in the approach space\label{line:construct_training_pair}.

        }

        Construct conditional training batch
        $\lambda = \{\lambda_{1}, \dots, \lambda_{m} \}$.

        Train $\mathcal{Q}_{\boldsymbol{\theta}}$ on $\lambda$ using Eq.~\eqref{eq:vae_loss}.
    }
\end{algorithm}

\subsection{Grasp Discriminator}

As \methodname{} is only trained on successful grasps, nothing prevents it from generating unsuccessful grasps between the modes \cite{mousavian20196}. Thus, we need a grasp discriminator $\mathcal{D}_{\boldsymbol{\psi}}(\text{S}=1 | \Vec{g} \in \matr{G}, \matr{O})$ to classify grasp success. $\mathcal{D}_{\boldsymbol{\psi}}$ is the same as in \cite{mousavian20196} and takes as input $\matr{O}$ concatenated with a grasp point cloud $\matr{K}\in \mathbb{R}^{6\times 3}$ and an additional binary point-wise feature $b \in \{0, 1\}$ that tells these two point clouds apart. The output of the discriminator is the probability a grasp will succeed. To construct $\matr{K}$ we use the same $\text{h}$ from Eq. \eqref{eq:rec_loss}. The grasp discriminator is trained on the binary cross-entropy loss. 

\subsection{Constrained Grasp Refinement}
\label{sec:refinement\methodname{}}

Previous works on unconstrained 6-\ac{dof} generative grasp sampling \cite{mousavian20196,murali20206} have highlighted that many of the grasps with low discriminator scores lie close to high-scoring grasps. Thus, doing local grasp refinements $\Vec{g} + \Delta \Vec{g}$ could significantly increase the probability of grasp success \ie{} $\prob{(\text{S}=1 | \Vec{g} + \Delta \Vec{g}, \matr{O})}>\prob{(\text{S}=1 | \Vec{g}, \matr{O})}$. Two prior works \cite{mousavian20196,murali20206} have proposed different grasp refinement methods. In \cite{mousavian20196}, the gradient of the discriminator is used to update the grasp pose  ($\Vec{g}_{t+1} = \Vec{g}_t +\kappa\partial \mathcal{D}_{\boldsymbol{\psi}}/\partial\Vec{g}_t$), while in \cite{murali20206} the Metropolis-Hasting algorithm is used to accept a randomly sampled refinement $\Delta \Vec{g}$ with the probability $\frac{\mathcal{D}_{\boldsymbol{\psi}}(\text{S}=1 | \Vec{g}+\Delta\Vec{g}, \matr{O})}{\mathcal{D}_{\boldsymbol{\psi}}(\text{S}=1 | \Vec{g}, \matr{O})}$. 

Unfortunately, neither of these refinement methods ensures that $\Vec{g}$ stays within $\matr{A}$. Therefore, we adapt the probability of success in the Metropolis-Hasting algorithm to be zero if the refined grasp $\Vec{g} + \Delta \Vec{g}$ is outside $\matr{A}$. Specifically, the probability of success changes to

\[
   \mathcal{D}_{\boldsymbol{\psi}}(\text{S}=1 | \Vec{g}+\Delta\Vec{g}, \matr{O})= 
\begin{cases}
    0,  \langle\Vec{v}_g, \Vec{v}_{\text{A}}\rangle\geq \cos{\alpha} & \\
    \mathcal{D}_{\boldsymbol{\psi}}(\text{S}=1 | \Vec{g}+\Delta\Vec{g}, \matr{O}),   \text{otherwise} &, 
\end{cases}
\]
where $\Vec{v}_g$ is the unit approach direction of the refined grasp $\Vec{g}+\Delta\Vec{g}$.

\subsection{\methodname{} Training Setup}

For training \methodname{}, we set $\text{L}=4$ and $\beta=\num{1e-2}$. \methodname{} is trained for 300 epochs with a batch size of 300 on three 2080Ti GPUs, starting with an initial learning rate of \num{2e-3} that is gradually annealed to 0. The total time for training \methodname{} is 57 hours.

\section{Dataset}

Thanks to the learning algorithm outlined in \secref{sec:training\methodname{}}, we can train \methodname{} on grasping datasets that do not contain conditional labels $\alpha$. Thus, we adapted the publicly available large-scale Acronym dataset \cite{eppner2021acronym} that contains 17.7M simulated parallel-jaw grasps on 8872 objects from ShapeNet \cite{chang2015shapenet} to our need. Of these objects, 112 were held out for the simulation experiments (\secref{sec:sim_exp}), while the rest were used for training (80\%) and validation (20\%).

\section{Experimental Evaluation}
\label{sec:Experiment}

\begin{table*}[ht]
    \centering
    \vspace{9pt}
    \begin{adjustbox}{max width=\linewidth}
    \setlength{\tabcolsep}{4pt}
         \begin{tabular}{lccccccc}
            \toprule
            & \multicolumn{5}{c}{\textbf{Angle Range (\text{\textdegree})}}\\
            \cmidrule(l){2-6}
            \textbf{Method}           & 30 & 45 & 60 & 75 & 90 & Avg\\ \midrule
            \methodname{}-nr       & 0.23 / 96.11\% & 0.26 / 98.70\%  & 0.27 / 99.61\%   & 0.26 / 99.88\%  & 0.23 / 99.96\%  & 0.25 / 98.85\%  \\ 
            \methodname{}-ur       & \textbf{0.45} / 67.70\% & \textbf{0.47} / 81.84\%  & \textbf{0.46} / 89.46\%   & \textbf{0.45} / 93.94\%  & \textbf{0.42} / 96.50\%  & \textbf{0.45} / 85.89\%  \\ 
            \methodname{}-cr       & 0.42 / \textbf{98.50\%} & 0.45 / \textbf{99.53\%}  & 0.45 / \textbf{99.89\%}   & 0.44 / \textbf{99.97\%}  & \textbf{0.42} / \textbf{99.99\%}  & 0.44 / \textbf{99.58\%}  \\ 
            \bottomrule
        \end{tabular}
    \end{adjustbox}
    \caption{Experiment results of \ac{auc} for the success-over-coverage rate and the kept ratio (\ac{auc}/Kept ratio) for \methodname{} w/wo grasp refinement. \methodname{}-nr denotes \methodname{} without refinement, \methodname{}-ur denotes \methodname{} with unconstrained refinement, and \methodname{}-cr denotes \methodname{} with constrained refinement.}
    \label{tab:sim_constrained_kept_ratio}
    \vspace{-9pt}
\end{table*}

With the experiments, we aim to answer the following questions:

1) How effective is \methodname{} in generating successful approach-constrained grasps?

2) To what extent does constrained grasp refinement enhance grasp quality?

3) How successful is \methodname{} at picking real objects in unconfined and confined spaces?

We address the above questions by conducting one simulation experiment using a free-floating gripper in Issac-Gym \cite{makoviychuk2021isaac} and two real-world experiments: one where objects are picked from a table and one where objects are picked from a shelf. In all experiments, a grasp is successful if the object is picked and remains within the gripper during a predefined movement.

\subsection{Simulation Experiments}
\label{sec:sim_exp}
\begin{figure}[ht]
    \centering
\includegraphics[width=0.8\linewidth]{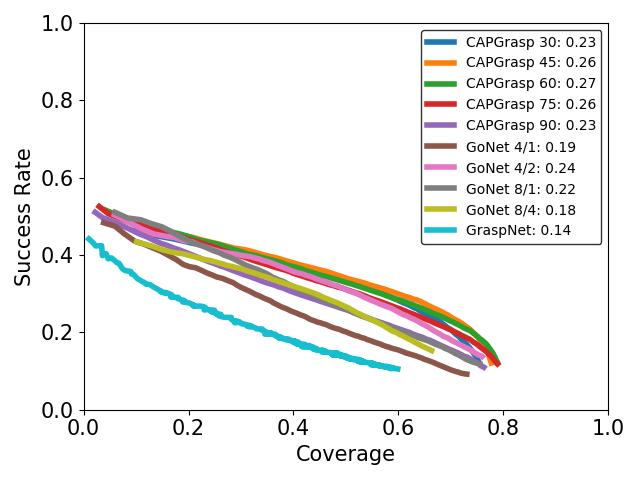}
    \vspace{-2mm}
    \caption{The success-over-coverage for \oldmethodname{} with different yaw/pitch resolutions, \methodname{} with different $\alpha$, and \graspnet{}. The \ac{auc} score for each method is visualized in the image legend.}
\label{fig:exp1_auc_cov_suc}
\vspace{-1pt}
\end{figure}

In the simulation environment, $\matr{O}$ is rendered using a simulated depth camera, and a virtual free-floating 7-\ac{dof} Franka Panda gripper is used for grasping. Gravity, initially off, is turned on after the gripper closes its fingers. To evaluate the stability of a grasp, the gripper performs a predefined movement consisting of a linear acceleration followed by an angular acceleration.

To address the first question, we evaluated all methods using the success-over-coverage rate metric from \cite{mousavian20196,weng2023gonet}. To calculate the coverage rate, we discretized the $\mathrm{SO}(3)$ space into 32 non-overlapping reference sectors. Each of these sectors contained a subset of the Acronym ground-truth grasps. All methods were allowed to sample 200 grasps for each sector. We deemed a ground-truth grasp covered by a generated grasp if the angle between their approach vectors is less than 10\textdegree and the translation distance is less than 2 cm.

We compared \methodname{} with different half angles $\alpha$ to \graspnet{} and four instances of \gonet{}, each with a different yaw/pitch resolution. The results, visualized in \figref{fig:exp1_auc_cov_suc}, demonstrate that \methodname{} is superior to all baselines, highlighting the benefits of a continuously constrained grasp sampler.

\begin{figure}[t]
    \centering
        \includegraphics[width=1.0\linewidth]{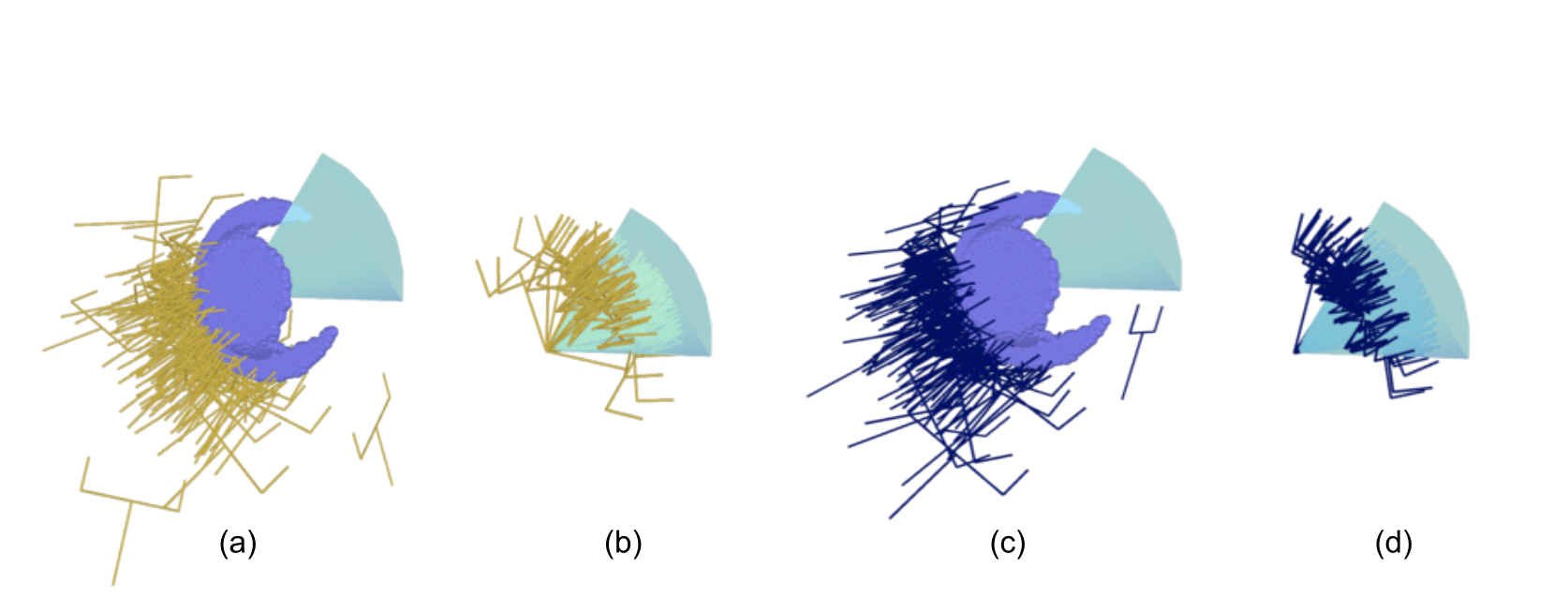}
    \caption{\methodname{} with unconstrained and constrained grasp refinement. Approach constraint is shown with a cyan cone ($\alpha=30$ \textdegree). (a) and (c) shows grasps after unconstrained and constrained refinement, respectively. (b) and (d) are the same grasps but translated to the origin of the spherical cone. Constrained refinement in (d) maintains grasps within constraints better than unconstrained refinement in (b).}
    \label{fig:refinement_process}
    \vspace{-1pt}
\end{figure}

To answer the second question, we studied the effect of unconstrained and constrained refinements on the success-over-coverage rate and on the ratio of grasps kept inside each sector. The results are presented in \tabref{tab:sim_constrained_kept_ratio}. Based on these results, we can conclude that refinement significantly improves grasp success, with similar gains across both refinement types. However, the kept ratio, which measures how many grasps are within $\matr{A}$, is consistently lower for the unconstrained than the constrained refinement, demonstrating the need for constrained refinement. For instance, with $\alpha$ lower than 60\textdegree, the kept ratio is between $10-30\%$ lower. The qualitative results on unconstrained and constrained refinement are shown in \figref{fig:refinement_process}.

For completeness, we also evaluated unconstrained refinement for \graspnet{} and \oldmethodname{}, achieving an \ac{auc} score of 0.24 and 0.48 and a kept ratio of 3.16\% and 64.64\%, respectively. We also performed constrained refinement of \oldmethodname{}, achieving a 0.44 AUC score and 96.70\% kept ratio, which, again, highlights the effectiveness of our constrained refinement method.

\begin{table*}[ht]
    \centering
    \vspace{9pt}
    \begin{adjustbox}{max width=\linewidth}
    \setlength{\tabcolsep}{2pt}
         \begin{tabular}{lccccccccc}
            \toprule
            \textbf{Experiment}& \multicolumn{9}{c}{\textbf{\# of successful grasps (Table-picking / Shelf-picking)}} \\
            \cmidrule(l){2-10} \cmidrule(l){6-10}
            Object & \graspnet{}-nr & \graspnet{}-ur & \gpd{}-ur & \oldmethodname{}-nr & \oldmethodname{}-cr & EGNet-1 & EGNet-2 & \methodname{}-nr & \methodname{}-cr \\ \midrule
            1. Bleach cleanser        & 3/2 & 5/3 & 4/3  & 4/2   & 5/3 & 3/1 & 5/1  & 5/3  & 5/4 \\ 
            2. Mustard bottle         & 1/1 & 3/2 & 3/1  & 3/3   & 3/2 & 3/2 & 3/1  & 2/2  & 5/3 \\ 
            3. Sugar box              & 5/2 & 5/3 & 4/2  & 5/2   & 5/3 & 4/3 & 4/4  & 5/3  & 5/4 \\ 
            4. Cracker box            & 4/0 & 3/1 & 2/0  & 5/1   & 5/3 & 2/0 & 2/1  & 5/2  & 5/3 \\
            5. Windex bottle          & 4/0 & 2/1 & 4/1  & 4/0   & 4/0 & 1/0 & 0/0  & 4/0  & 3/1  \\
            6. Metal bowl             & 5/2 & 5/5 & 5/5  & 5/4   & 5/5 & 4/4 & 4/5  & 5/5  & 5/5\\
            7. Tennis ball            & 1/1 & 4/3 & 0/1  & 4/2   & 5/3 & 3/3 & 5/2  & 5/5  & 4/3 \\
            8. Metal mug              & 5/1 & 4/1 & 0/2  & 5/3   & 4/4 & 4/0 & 5/2  & 5/3  & 5/2 \\
            9. Banana                 & 0/0 & 0/0 & 0/0  & 3/0   & 2/0 & 2/0 & 2/0  & 4/0  & 4/0 \\
            10. Spring clamp          & 0/0 & 2/0 & 2/0  & 2/0   & 2/0 & 0/0 & 3/0  & 3/0  & 4/0 \\
                        \midrule 
            Avg. success rate  &56\%/18\%& 66\%/38\% &48\%/30\% & 80\%/34\%  & 80\%/46\% & 52\%/26\% & 66\%/32\% & 86\%/46\% & \textbf{90\%}/\textbf{50\%} \\
            Kept ratio  &-\%/-\%& 16\%/22\% &8\%/33\% & \textbf{99\%}/\textbf{99\%}  & \textbf{99\%}/\textbf{99\%} & -\%/-\% & 33\%/51\% & 98\%/97\% & 98\%/\textbf{99\%}
            \\
            
            \bottomrule
        \end{tabular}
    \end{adjustbox}
    \caption{Experimental results for table-picking and shelf-picking. For each object and experiment, we report the number of successful grasps out of maximum 5. The suffixes -nr, -ur, and -cr denote no refinement, unconstrained refinement, and constrained refinement, respectively. EGNet-1 represents pre-trained \edgegrasp{} without filtering. EGNet-2 represents pre-trained \edgegrasp{} with filtering. }
    \label{tab:real_exp}
    \vspace{-9pt}
    \end{table*}
    
\subsection{Real-world Experiments}

To answer the third question, we conducted two real-world experiments, table- and shelf-picking. The experiments are carried out on a Franka Emika Panda robot. We use a KinectV2 to capture $\matr{O}$ and an Aruco Marker \cite{arucomarker} for extrinsic camera calibration. The camera was placed to view the scene from the side. We used the same ten daily-used objects as in \cite{weng2023gonet}. These objects were placed in the workspace in 5 orientations: 0\textdegree, 72\textdegree, 144\textdegree, 216\textdegree, and 288\textdegree. 

We compared \methodname{} to four different baselines: the non-learning-based grasp sampler in GPD \cite{ten2017grasp}; the learning-based grasp sampler \graspnet{} \cite{mousavian20196}; the constrained but discretized learning-based grasp sampler \gonet{} \cite{weng2023gonet}; and the \sota{} SE(3)-equivariant grasp detector \edgegrasp{} \cite{huang2023edge}. All methods were assessed based on grasp success rate and kept ratio. For each object orientation, \edgegrasp{} sampled and evaluated 2000 grasps, out of which the top 200 were retained. The other methods sampled and scored 200 grasps directly, with and without performing grasp refinement with the discriminator. For all methods, the first feasible grasp among the top 10 highest-scoring grasps was executed.

\begin{figure}[t]
    \centering
\includegraphics[width=0.8\linewidth]{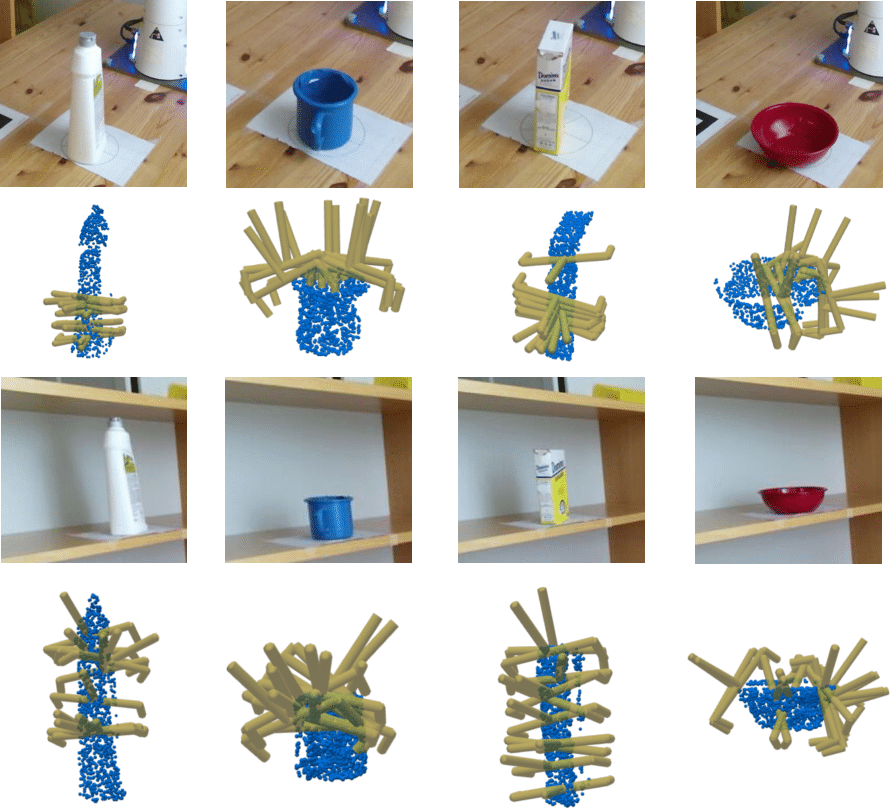}
    \caption{Top 10 grasps by \methodname{} for table and shelf-picking tasks after refinement.}
\label{fig:qualitative_real_comparison}
\vspace{-1pt}
\end{figure}

In the table-picking experiment, the approach directions were consistent with \cite{weng2023gonet}. Specifically, we used the second principal component of the object point cloud and the top-down direction as the approach directions. For the shelf-picking task, we chose five cone constraints. These cones effectively represented the collision-free space around the object as depicted in \figref{fig:confine_space_bowl}. We fix $\alpha$ to 30\textdegree, as this setting achieved the highest grasp success rate in simulation when the 10 highest-scoring grasps were executed.

The experimental results are summarized in Table~\ref{tab:real_exp}.  Overall, the average grasp success rate for \methodname{} is 10\%--38\% higher than the baselines on the table-picking task and 4\%--32\% higher in the shelf-picking task. The reason behind the higher success rates is that the continuous constraint in \methodname{} is more flexible than the discretized constraints in \oldmethodname{}, especially in confined environments. Similar to the simulation experiments, we can once again see that constrained refinement benefits both \oldmethodname{} and \methodname{} with an average improvement of 0\%--14\%  for grasp success rate while maintaining a high kept ratio. A detailed breakdown of the grasp success rate for each object is shown in Table~\ref{tab:real_exp}. These breakdown results show that the baseline methods struggle with large and flat objects, whereas \methodname{} excels across all objects. The qualitative results on the 10 highest-scoring grasps by \methodname{} on 4 different objects are visualized in \figref{fig:qualitative_real_comparison}. 

Another observation is that learning-based methods generally outperform the non-learning-based GPD method. The main reason GPD is inferior to learning-based methods is that the grasps it samples are less diverse, with few to none on the occluded regions of the object, making it oftentimes impossible to find kinematically feasible paths to any of the sampled grasps. This also happens to \edgegrasp{} \cite{huang2023edge} that relies on non-learning-based sampling.

\section{Conclusion}

This paper addresses the problem of continuous approach-constrained grasp sampling. The proposed solution includes \methodname{}, a novel continuous approach-constrained \equiv{} generative grasp sampler, a new algorithm for training continuous conditioned grasp samplers that circumvents the need for large offline datasets, and a grasp refinement technique that improves sampled grasps while respecting approach constraints. Together, these innovations improved grasp success rates by 4-38\% compared to existing unconstrained and non-continuous constrained methods.  

Despite the advancement, our work has two limitations, both stemming from the constraints. The first limitation is that $\matr{A}$ is selected manually or through some simple-to-compute geometrical features. For instance, in the shelf-picking task, we manually set $\Vec{v}_A$ and $\alpha$ such that the generated grasps would have a low chance of colliding with the environment. The second limitation is that we can only pose isotropic cone constraints but not non-isotropic elliptical cone constraints. The benefit of elliptical cone constraints is that only one could replace the current five cone constraints used to grasp objects from a shelf. These two limitations are exciting areas for future research.

Overall, the presented solutions generalize orientation constraint handling to continuous subregions of SO(3). We envision that continuously constrained grasp samplers could play a key role in completing robotic manipulation tasks where grasping specific locations is a prerequisite for task success, such as when handing objects to humans. Nevertheless, to use the proposed solutions to solve such tasks, the constraint should be automatically inferred based on the context, perhaps using the recent advancements of Large Vision Language models, which is a challenging but exciting future work direction.


%



\section*{Acknowledgment}

We would like to show our gratitude to the Swedish Research Council, Knut and Alice Wallenberg Foundation, and Horizon2020 ERC BIRD.

\ifCLASSOPTIONcaptionsoff
  \newpage
\fi

\bibliographystyle{IEEEtran}
\bibliography{ref}

\end{document}